\documentclass[a4paper,fleqn]{cas-sc}
\newcommand\figref{Fig.~\ref}

\usepackage[round]{natbib}
\usepackage{mathtools}
\usepackage{amsmath,amsfonts,amsthm,bm} 
\usepackage{textcomp}
\usepackage{siunitx}
\usepackage{xcolor}
\usepackage{booktabs}
\usepackage{cases}
\usepackage[normalem]{ulem}
\usepackage{subcaption}
\usepackage{caption}
\usepackage[switch,columnwise]{lineno}
\usepackage{colortbl}  
\definecolor{lightyellow}{RGB}{255, 255, 170}
\definecolor{lightygray}{RGB}{211,211,211}
\usepackage{letltxmacro}
\LetLtxMacro{\originaleqref}{\eqref}
\renewcommand{\eqref}{Eq.~\originaleqref}

\def\tsc#1{\csdef{#1}{\textsc{\lowercase{#1}}\xspace}}
\tsc{WGM}
\tsc{QE}
\tsc{EP}
\tsc{PMS}
\tsc{BEC}
\tsc{DE}

\begin{document}
\let\WriteBookmarks\relax
\def\floatpagepagefraction{1}
\def\textpagefraction{.001}

\shorttitle{Deformation-Aware GNN for Cancer Site Prediction}

\shortauthors{Kyunghyun Lee et~al.}

\title [mode = title]{Real-time prediction of breast cancer sites using deformation-aware graph neural network}

\author[1]{Kyunghyun Lee}
\cormark[1]
\affiliation[1]{organization={Yonsei University, School of Mathematics and Computing
(Computational Science and Engineering)},
    city={Seoul},
    postcode={03722},
    country={Republic of Korea}}

\author[1]{Yong-Min Shin}
\cormark[1]
\cortext[cofirst]{Kyunghyun Lee and Yong-Min Shin contributed equally to this work.}

\author[2]{Minwoo Shin}
\affiliation[2]{organization={Yonsei University, Division of Software},
    city={Wonju},
    postcode={26493},
    country={Republic of Korea}}

\author[3]{Jihun Kim}
\affiliation[3]{organization={Yonsei University College of Medicine, Department of Radiation Oncology},
    city={Seoul},
    postcode={06273},
    country={Republic of Korea}}

\author[4]{Sunghwan Lim}
\affiliation[4]{organization={Korea Institute of Science and Technology, Bionics Research Center},
    city={Seoul},
    postcode={02792},
    country={Republic of Korea}}

\author[1, 5]{Won-Yong Shin}
\cormark[2]
\ead{wy.shin@yonsei.ac.kr}

\author[1, 5]{Kyungho Yoon}
\affiliation[5]{Graduate School of Artificial Intelligence, Pohang University of Science and Technology (POSTECH), Pohang, 37673, Republic of Korea}
\cormark[2]
\ead{yoonkh@yonsei.ac.kr}
\cortext[cor1]{Corresponding author}

\begin{abstract}
Early diagnosis of breast cancer is crucial, enabling the establishment of appropriate treatment plans and markedly enhancing patient prognosis. While direct magnetic resonance imaging (MRI)-guided biopsy demonstrates promising performance in detecting cancer lesions, its practical application is limited by prolonged procedure times and high costs. To overcome these issues, an indirect MRI-guided biopsy that allows the procedure to be performed outside of the MRI room has been proposed, but it still faces challenges in creating an accurate real-time deformable breast model. In our study, we tackled this issue by developing a graph neural network (GNN)-based model capable of accurately predicting deformed breast cancer sites in real time during biopsy procedures. An individual-specific finite element (FE) model was developed by incorporating magnetic resonance (MR) image-derived structural information of the breast and tumor to simulate deformation behaviors. A GNN model was then employed, designed to process surface displacement and distance-based graph data, enabling accurate prediction of overall tissue displacement, including the deformation of the tumor region. The model was validated using phantom and real patient datasets, achieving an accuracy within 0.2 millimeters (mm) for cancer node displacement (root mean squared error, RMSE) and a dice similarity coefficient (DSC) of 0.977 for spatial overlap with actual cancerous regions. Additionally, the model enabled real-time inference and achieved a speed-up of over 4,000 times in computational cost compared to conventional FE simulations. The proposed deformation-aware GNN model offers a promising solution for real-time tumor displacement prediction in breast biopsy, with high accuracy and real-time capability. Its integration with clinical procedures could significantly enhance the precision and efficiency of breast cancer diagnosis.

\end{abstract}


\begin{highlights}
\item This study presents a real-time tumor deformation prediction model using GNN.
\item The GNN model showed high predictive accuracy and real-time capability.
\item The approach significantly improves the precision and efficiency of breast biopsy procedures.
\end{highlights}


\maketitle

\section{Introduction}
Breast cancer is the most prevalent cancer and fatal cause of cancer-related deaths among women \citep{bray2018global}. Early diagnosis is critical to improving survival rates, as detecting cancer at its early stages enables more effective treatment options, which can significantly reduce mortality \citep{smith2003american}. A core needle biopsy, often assisted by mammography and ultrasound imaging, is the most commonly utilized and preferred technique for breast cancer diagnosis \citep{BbioMayo}. However, both ultrasound- and mammography-guided methods have some limitations, such as depending heavily on the skill of the operator, reduced accuracy, and even the possibility of missing certain types of breast lesions \citep{o2010image, helbich2004stereotactic, schueller2008accuracy, mann2019breast}.

Meanwhile, breast magnetic resonance imaging (MRI) is widely recognized for its superior sensitivity and specificity in detecting suspicious cancer lesions, with studies showing that more than half of these lesions are visible only on MRI \citep{berg2012}. To leverage this advantage, MRI-guided breast biopsy, performed directly inside an MRI suite where real-time scans are used to accurately locate and target suspicious lesions, has been successfully validated in clinical practice \citep{price2013magnetic}. Nevertheless, in clinical practice, MRI-guided biopsy must be performed within an MRI suite, where scheduling is highly constrained due to limited scanner availability and long waiting lists. As a result, only a small fraction of eligible patients can undergo this procedure, despite its diagnostic accuracy. The prolonged procedural time, high operational cost, and logistical difficulties significantly limit its accessibility and scalability in routine clinical settings \citep{saslow2007}.

To overcome these limitations, an alternative approach using an indirect MRI-guided breast biopsy with machine vision techniques has been proposed \citep{IndirectMRI2024}, allowing the biopsy to be performed outside the MRI suite. This method estimates the positions of targets within the breast by using pre-acquired magnetic resonance (MR) images and performing real-time deformable registration between a deformable breast model and real-time shape-sensing data. By reducing procedure time and cost, this approach aims to make indirect MRI-guided biopsies more accessible for routine clinical use, which could help more patients benefit from MRI’s high precision in cancer detection and lead to increased opportunities for early diagnosis and treatment. Although the study demonstrated partial feasibility, a major limitation remains: the absence of an accurate, real-time deformable breast model. Current methods rely solely on static, pre-acquired MRI scans and thus cannot capture the complex deformation behavior of breast tissue. This limitation is further compounded by the tissue’s inherently high elasticity and deformability, which often cause substantial lesion displacement during compression or external manipulation. Consequently, targeting accuracy is significantly reduced in clinical practice. To enable reliable indirect MRI-guided biopsy, it is therefore critical to develop a deformable model capable of real-time, patient-specific response.

Finite element (FE) analysis has established itself as a fundamental tool in science and engineering research, enabling detailed modeling of complex systems and facilitating the analysis of deformation behaviors under various conditions \citep{ghahramani2024,eggermont2024, baek2016, yoon2015}. Its versatility has led to widespread applications in various biomedical analyses, ranging from tissue deformation studies to the simulation of physiological processes \citep{sharifzadeh2020,sabet2021,kim2023laser, karimi2021}. Despite its advantages, FE analysis faces significant limitations due to its high computational cost, which often hinders real-time operation and restricts its applicability. Various model order reduction techniques, such as proper orthogonal decomposition, have been proposed to address this issue; however, while effective in linear or weakly nonlinear regimes, these methods often struggle with the strongly nonlinear and heterogeneous mechanical behavior observed in complex systems \citep{quarteroni2015reduced, benner2015survey}.

Recently, a hybrid approach that integrates high-fidelity numerical simulations with machine learning (ML) has gained attention as a promising method to overcome this limitation, offering both real-time performance and high predictive accuracy in modeling complex systems \citep{liang2018, shin2024, choi2022, seo2024multi, shin2023multi, Zhao2022Graphformer}. Although ML-based approaches can utilize the abundance of software tools to train ML models alongside their general applicability to learn from data, sufficient curation and generation of training data acts as a key aspect in the overall process \citep{zelaya2019towards, chen2024automated, bhardwaj2024machine}. While an FE mesh is typically used to model the tissue/organ of interest, many studies have employed graph neural networks (GNNs) as their main architecture to learn from the mesh data~\citep{Pfaff2021graphmeshnet, Gladstone2024MeshbasedGS}. Unlike conventional deep learning models such as convolutional neural networks (CNNs), which are optimized for regular grid-like data structures, we require a deep learning architecture fit for learning from irregular non-Euclidean data, e.g., FE meshes. The most natural data structure to express such FE meshes are networks or graphs, which makes GNN models the best architecture choice. The ability of GNNs to naturally encode both nodal features (e.g., displacement vectors) and structural information from the mesh makes them particularly effective in capturing the complex relationships within the system \citep{pagan2022, dalton2022, jiang2023, Salehi2022PhysGNN, sen2024physics}.

Leveraging the structural advantages of GNN, we developed a real-time model that predicts the deformed cancer location during breast biopsy, based on a GNN trained with FE analysis results. The FE model was constructed to simulate individual-specific deformation behaviors by incorporating the structural shapes of the breast and tumor, as characterized by the respective MRI data. Then, we employ a GNN model specifically designed to incorporate the surface displacement and distance-based graph information to predict the deformation of all nodes, effectively learning the underlying physical phenomenon. Additionally, by using surface displacement information, which can be measured in real-clinical environments instead of the difficult-to-measure force as an input feature, the model demonstrates its practical applicability. The model was validated through phantom and real patient datasets, demonstrating the capability to predict cancer node displacement with an accuracy within 0.2 mm in terms of root mean squared error (RMSE), and spatial overlap between the predicted and actual cancerous regions, with a dice similariy coefficient (DSC) of 0.977. Moreover, the model enabled real-time inference and achieved a speed-up of over 4,000 times in computational cost compared to conventional FE simulation. Our main contributions are summarized as follows:
\begin{itemize}
    \item GNN encodes the structural feature of FE model, enabling real-time, high-accuracy prediction.
    \item MRI data was used to build a FE model reflecting individual-specific breast and tumor features.
    \item The model enhances its applicability by using  a measurable feature in clinical settings. 
    \item The model was validated using real patient datasets, showing its robustness and real-world applicability.
\end{itemize}

The remainder of this paper is organized as follows. Section 2 describes the generation of training data using subject-specific finite element simulations. Section 3 presents the architecture and training process of the proposed graph neural network model. Section 4 reports the experimental results and performance evaluation on both phantom and patient datasets. Finally, the paper concludes with a discussion and summary of the findings.

\section{Training data generation}
In this section, the process of generating the training dataset through FE analysis is introduced, including the FE formulation, the construction of the FE model using MRI data, and the dataset generation through randomly applied surface forces.

\subsection{FE formulation}
The deformation of all solids and structures is based on continuum mechanics. The nonlinear FE formulation, known as the total Lagrangian formulation for element $m$, is derived from continuum mechanics as follows   \citep{bathe2006, yoon2014, jun2018}:
\begin{equation}
\begin{split}
\int_{\prescript{0}{}{V}} \delta \prescript{}{0}{e_{ij}} \prescript{}{0}{C_{ijrs}}  \prescript{}{0}{e_{rs}} \, d \prescript{0}{}{V}
+ \int_{\prescript{0}{}{V}} \prescript{t}{0}{S_{ij}} \delta \prescript{}{0}{\eta_{ij}} \, d \prescript{0}{}{V} \\
= \prescript{t+\Delta t}{}{R} - \int_{\prescript{0}{}{V}} \prescript{t}{0}{S_{ij}} \delta \prescript{}{0}{e_{ij}} \, d \prescript{0}{}{V}
\end{split}
    \label{eq:leq}
\end{equation}
where $\prescript{0}{}{V}$ is volume of initial configuration, $\prescript{}{0}{e_{ij}}$ and $\delta$$\prescript{}{0}{e_{ij}}$ respectively represent a linear part of incremental Green-Lagrange (GL) strain and its virtual component, $\prescript{}{0}{C_{ijrs}}$ is material law tensor, $\delta$$\prescript{}{0}{\eta_{ij}}$ is a virtual component of nonlinear incremental GL strain, $\prescript{t}{0}{S_{ij}}$ is a second Piola-Kirchhoff stress, and $\prescript{t+\Delta t}{}{R}$ is external virtual work at time $t+\Delta t$.

After applying the standard finite element discretization procedure \citep{yoon2014, jun2018}, the incremental strains and their virtual components can be expressed in terms of nodal displacements and its virtual component ($\textbf{U}$ and $\Delta \textbf{U}$) as

\begin{equation}
{}_{0}e^{(m)}_{ij} = \textbf{B}^{(m)}_{ij} \Delta \textbf{U}, 
\quad 
\delta_{0}e^{(m)}_{ij} = \textbf{B}^{(m)}_{ij} \, \delta \Delta \textbf{U},
\quad
\delta{}_{0}\eta^{(m)}_{ij} = \delta \Delta \textbf{U}^T \textbf{N}^{(m)}_{ij} \, \Delta \textbf{U},
\end{equation}
where $\textbf{B}^{(m)}_{ij}$ and $\textbf{N}^{(m)}_{ij}$ are the linear and nonlinear strain-displacement matrices of element m, respectively.

Substituting these relations into ~\eqref{eq:leq} and assembling the element contributions yield the following matrix form at the element level:
\begin{align}
\delta \Delta \textbf{U}^T\sum_{m} \Bigg[
& \int_{{}^{0}V^{(m)}} 
\big( \textbf{B}^{(m)}_{ij} \big)^{T} \, {}_{0}C^{(m)}_{ijkl} \, \textbf{B}^{(m)}_{kl} \, dV^{(m)} \notag  + \int_{{}^{0}V^{(m)}} 
\textbf{N}^{(m)}_{ij}  \, {}^{t}_{0}S^{(m)}_{ij} \, dV^{(m)}
\Bigg] \Delta \textbf{U} \notag \\
& = \delta \Delta \textbf{U}^T \bigg[ {}^{t+\Delta t} \textbf{R} 
- \sum_{m} \int_{{}^{0}V^{(m)}} 
\big( \textbf{B}^{(m)}_{ij} \big)^{T} \, {}^{t}_{0}S^{(m)}_{ij} \, dV^{(m)}\bigg]
\end{align}
in which the first term represents the linear component of stiffness, the second term accounts for the nonlinear component of stiffness, and the right-hand side corresponds to the external and internal nodal force imbalance. For clarity, the element stiffness matrices and internal force vector are defined as
\begin{equation}
\textbf{K}^{(m)}_{L} 
= \int_{{}^{0}V^{(m)}} 
\big( \textbf{B}^{(m)}_{ij} \big)^{T} \, {}_{0}C^{(m)}_{ijkl} \, \textbf{B}^{(m)}_{kl} \, dV^{(m)},
\end{equation}
\begin{equation}
\textbf{K}^{(m)}_{NL} 
= \int_{{}^{0}V^{(m)}} 
\textbf{N}^{(m)}_{ij}  \, {}^{t}_{0}S^{(m)}_{ij} \, dV^{(m)},
\end{equation}
\begin{equation}
{}^{t}_{0}\textbf{F}^{(m)} 
= \int_{{}^{0}V^{(m)}} 
\big( \textbf{B}^{(m)}_{ij} \big)^{T} \, {}^{t}_{0}S^{(m)}_{ij} \, dV^{(m)}.
\end{equation}
By assembling the contributions of all elements, the global tangential stiffness matrix (${}^{t}_{0}\textbf{K}$) and internal force vector (${}^{t}_{0}\textbf{F}$) are obtained as

\begin{equation}
{}^{t}_{0}\textbf{K} = \sum_{m} \Big( \textbf{K}^{(m)}_{L} + \textbf{K}^{(m)}_{NL} \Big), 
\qquad
{}^{t}_{0}\textbf{F} = \sum_{m} {}^{t}_{0}\textbf{F}^{(m)}.
\end{equation}
This leads to the incremental equilibrium equation,
\begin{equation}
\label{eq:iee}
\prescript{t}{0}{\textbf{K}}\Delta \textbf{U}=\prescript{t+\Delta t}{}{\textbf{R}} - \prescript{t}{0}{\textbf{F}}
\end{equation}

~\eqref{eq:iee} is iteratively solved using the Newton-Raphson method until convergence, allowing for the consideration of both geometric and material nonlinear effects in the analysis. The entire FE procedure was automated and executed using ANSYS Mechanical with batch files, enabling the analysis of multiple cases efficiently \citep{thompson2017ansys}.

\subsection{FE modeling through MRI data}
A breast phantom, named P1 herein, containing an internal tumor mass (breast model 3401, GPI Anatomicals, IL) is prepared. To obtain the structural information of the phantom, a $T_1$-weighted high-resolution image (field of view 15 $\times$ 15 cm$^2$, slice thickness 1.0 mm, image matrix 160 $\times$ 160, number of slices 150, repetition time 6.193 ms, echo time 2.303 ms, flip angle 10$^\circ$) is obtained by the 3 Tesla MR scanner (Achieva, Philips, Netherlands). 

Due to their distinct signal intensities in $T_1$-weighted images, normal breast and cancer tissue regions are first segmented using simple thresholding. This segmentation allows us to explicitly separate cancerous and normal regions in the reconstructed 3D geometry, which was exported in OBJ format. During mesh generation with ANSYS SpaceClaim (2021 R1), the elements are automatically tagged according to the segmented regions. Consequently, two distinct finite element groups are created: one corresponding to normal tissue and the other to cancer tissue. The target quality of the mesh is set to 0.05, resulting in an average surface area of $2.2987\times 10^{-5} m^2$ and a minimum edge length of $1.6802\times 10^{-3}m$. These parameters are confirmed to provide sufficient accuracy through solution convergence testing. 10-node tetrahedral quadratic elements (named Tet10 in ANSYS) are assigned to the meshes for constructing the FE model, resulting in 9,219 elements, 15,864 nodes, and 47,592 degrees of freedom (DOFs). To account for the nonlinearity of soft tissue deformation, the Mooney-Rivlin two-parameter model was used. The material constants ($C_{10}$ and $C_{01}$) were adopted from Yin et al. \citep{yin2004imageparser}, which estimated parameters based on biomechanical modeling of breast tissues. Specifically, values of $C_{10} = 2{,}000$ Pa and $C_{01} = 1{,}333$ Pa were assigned to normal breast tissue, while $C_{10} = 10{,}000$ Pa and $C_{01} = 6{,}667$ Pa were used for cancer tissue. In the FE formulation, these parameters enter the material law tensor $C_{ijkl}$, ensuring that the distinction between cancer and normal tissue is consistently preserved from segmentation to FE analysis. The DOFs at the nodes on the bottom surface are fixed and used as boundary conditions (BCs). Using the connectivity information of the nodes consisting the FEs, the local stiffness matrix of each tetrahedral element is assembled to obtain the tangential stiffness matrix $\prescript{t}{0}{\textbf{K}}$ in ~\eqref{eq:iee}, which is updated at each Newton-Raphson iteration.

\begin{figure}[ht!]
    \centering
\includegraphics[width=0.45\textwidth]{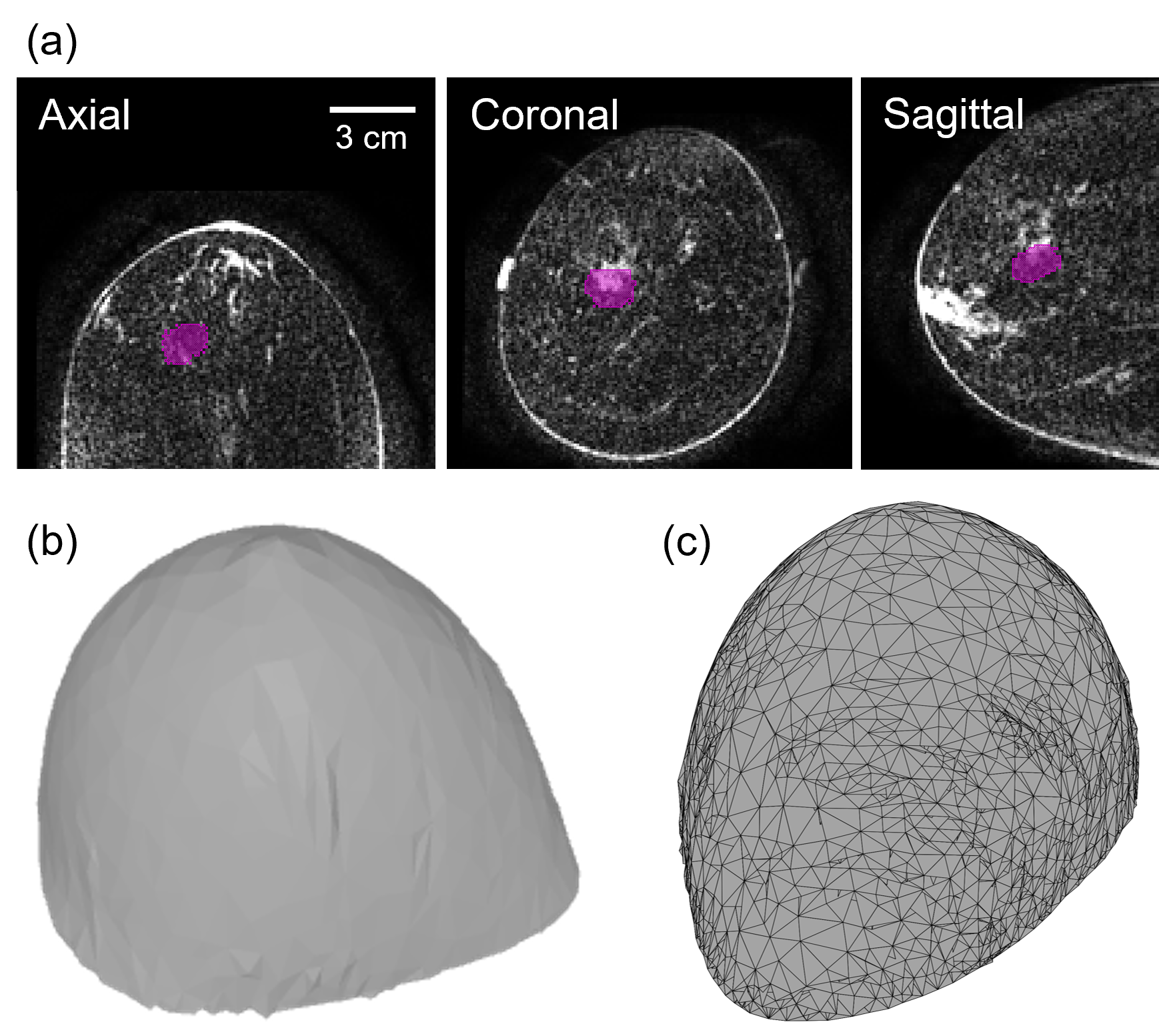}
\caption{Example of a finite element model constructed from an MR image. (a) Triplanar view of a breast MR image. The purple region indicates the cancerous area. (b) 3D surface data extracted from the MR image. (c) Finite element mesh representation.}
    \label{fig:fig_modeling}
\end{figure}

In addition to the breast phantom, MR images from three breast cancer patients (named H1, H2, and H3 herein), sourced from a publicly available dataset \citep{saha2018}, were used for further validation. The datasets included dynamic contrast-enhanced (DCE) MR images and corresponding segmented cancer images, which were annotated by radiologists to depict the cancerous regions for each patient. To construct the FE model, these MR images were processed in the same way as the P1 data, resulting  in 9,744 elements, 16,985 nodes, and 50,955 DOFs for H1; 8,562 elements, 15,064 nodes, and 45,192 DOFs for H2; and 9,810 elements, 16,935 nodes, and 50,805 DOFs for H3, respectively. The corresponding MRI images, 3D reconstruction, and meshing results for H1 are shown in \figref{fig:fig_modeling}.

\subsection{Dataset generation via FE analysis}
Our training dataset, consisting of the displacement at surface nodes ($\bold{U}_s$) and the displacement at all nodes ($\bold{U}$), was generated by performing the FE analysis using the proposed FE model. In a standard FE system, the input is a given force and the output is the displacement at all nodes. However, in this study, instead of using force as an input—which is not directly measurable in a clinical setting—the displacement at top surface nodes, which can be captured by a depth camera, was used as the input data.

To cover various deformation patterns, FE analysis were performed for a total of 3,000 loading conditions by applying forces with 300 different randomly chosen magnitudes, directions, and distribution ranges at 10 different loading locations. Among these locations, five were selected to be evenly distributed around the cancer tissue, while the remaining five were positioned to ensure a balanced distribution across the entire region. To achieve deformations well matched to real-world conditions, distributed loads were used instead of point loads. The load distribution was defined as a randomly sized spherical region with a radius ranging from 0.02 m to 0.032 m. To consider only situations where compressive deformation occurs, the direction of each force is randomly assigned with the constraint that the vertical component ({\it i.e.}, the y-direction) is set to be negative. The magnitude of the force is also randomly assigned within a range of 120 N to 240 N. In summary, the randomized loading configuration for each case is as follows:
\begin{itemize}
  \item \textbf{Location}: 10 total locations (5 near the cancer region and 5 across the general area),
\item \textbf{Direction}: Randomized with the constraint that the vertical component is negative,
\item \textbf{Magnitude}: Randomized uniformly between 120 N and 240 N,
\item \textbf{Distribution area}: Spherical region with radius randomly chosen between 0.02 m and 0.032 m.
\end{itemize}

These external force vectors were incorporated into the FE simulation as the input term ($\prescript{t+\Delta t}{}{\textbf{R}}$) in the incremental equilibrium equation. The nonlinear system was solved using the Newton-Raphson method, where at each iteration, the incremental displacement vector $\Delta \mathbf{U}$ was computed and accumulated until convergence was reached. This yielded the final nonlinear displacement vector $\mathbf{U}$ for all nodes. From the simulation result, displacements at the surface nodes ({\it i.e.}, $\mathbf{U}_s$) were obtained by retaining the displacement values at top surface nodes, while setting the displacement values at all other nodes to zero. The three dimensional displacement components  $\delta_x$, $\delta_y$, $\delta_z$ from both $\mathbf{U}_s$ and $\mathbf{U}$ at each node were then used as the input and output node features for the GNN model. The total of 3,000 cases were split into 2,400, 300, and 300 samples for training, validation, and testing, respectively.

In addition, to specifically evaluate the model's performance in localizing cancer, displacement vectors at cancer nodes were separately extracted from $\mathbf{U}$ based on the pre-tagged segmentation information, and used for detailed analysis.

\section{Training the GNN model}

GNN is a type of neural network designed for processing irregular non-Euclidean data, commonly represented as graphs~\citep{zhou2020graph}. Given a graph, GNNs can extract the useful features that encode both the nodal information and graph structural information through iterative message-passing mechanisms. As GNNs are also built upon neural networks, they also provide the flexibility where one can train them to provide features to solve a variety of downstream graph-related tasks. In this section, we first describe the formal description of the graphs that we have constructed. Then, we provide a detailed description of our GNN model architecture for deformation prediction. In our framework, we leverage GNNs to learn the mapping from observable surface deformation to full volumetric deformation of soft tissue. Rather than explicitly modeling the physical process of tissue deformation within the network, we use GNNs to learn a data-driven relationship between surface displacement ($\mathbf{U}_s$) and the corresponding internal displacement ($\mathbf{U}$), which includes regions affected by cancer. In this section, we first describe the definition and construction of the graphs used to represent the breast tissue deformation. Then, we provide a detailed description of our GNN model architecture for deformation prediction.

\begin{figure*}[t!]
    \centering
    \begin{subfigure}[b]{0.35\textwidth}
        \includegraphics[width=\textwidth]{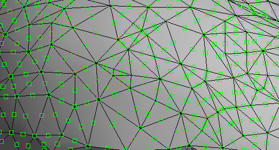}
        \caption{Node condition}
        \label{fig:nodeCond}
    \end{subfigure}
    \hspace{2mm}
    \begin{subfigure}[b]{0.22\textwidth}
        \includegraphics[width=\textwidth]{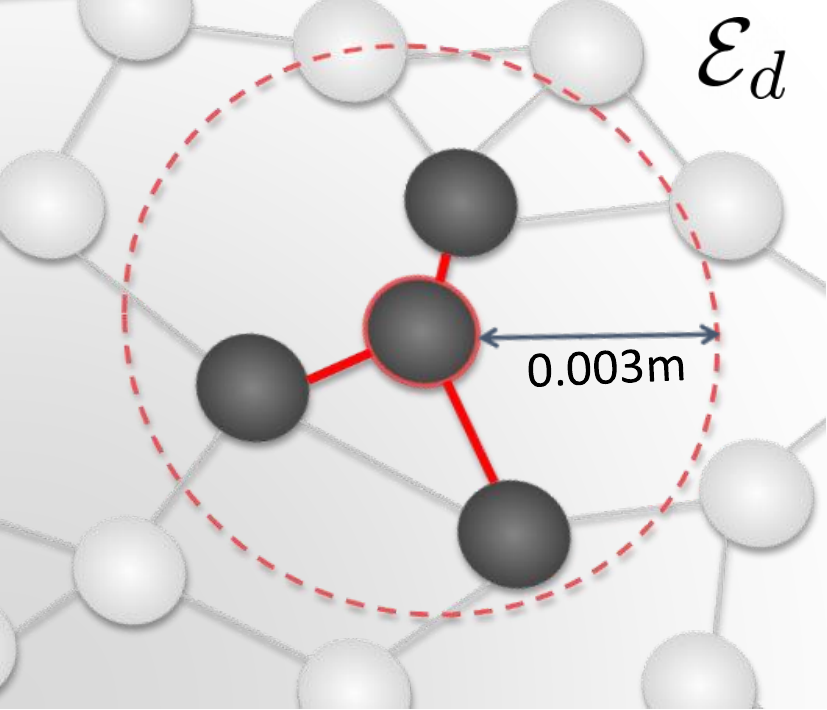}
        \caption{Condition for $\mathcal{E}_d$}
        \label{fig:edgeCond}
    \end{subfigure}
    \hspace{2mm}
    \begin{subfigure}[b]{0.22\textwidth}
        \includegraphics[width=\textwidth]{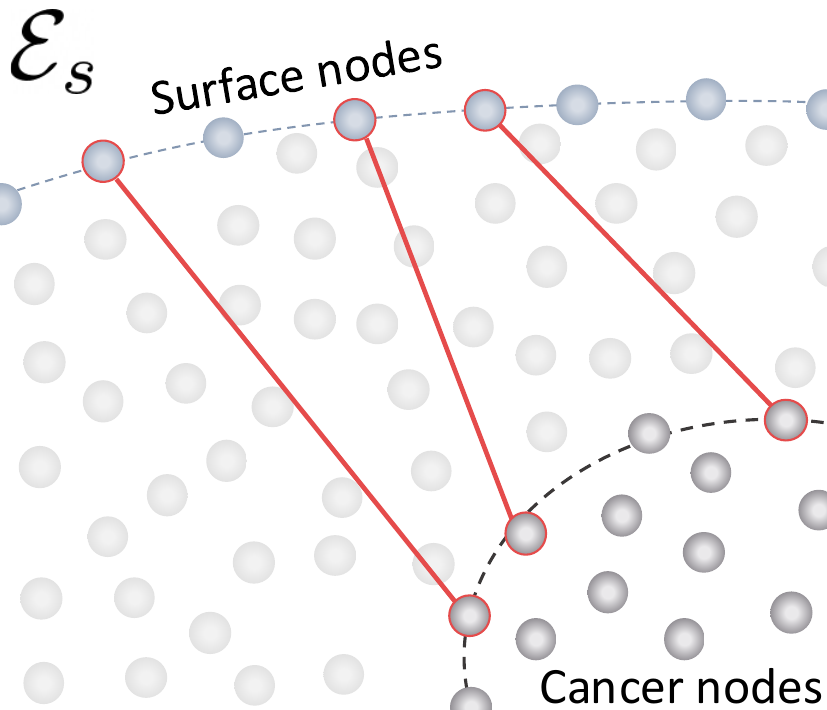}
        \caption{Condition for $\mathcal{E}_s$}
        \label{fig:edgeCondStruct}
    \end{subfigure}
    \caption{Graph construction process:  
(a) Graph nodes are created to match the nodes of the FE model. (b) Distance-based edge construction ($\mathcal{E}_d$): Nodes within 0.003 m of each other are connected to capture local spatial relationships.  
(c) Structured edge augmentation ($\mathcal{E}_s$): A subset of surface nodes and cancer surface nodes are directly connected to enhance long-range information propagation.}
    \label{fig:graphStruct}
\end{figure*}

\subsection{Graph construction} \label{subsection:graphconstruction}

Formally, a graph is defined as a tuple of sets $G = (\mathcal{V}, \mathcal{E})$, where $\mathcal{V}$ is the set of nodes and $\mathcal{E} \subseteq \mathcal{V} \times \mathcal{V}$ is the set of edges. In our framework, $\mathcal{V}$ directly inherits the nodes of the FE model used in the mechanical simulation, as illustrated in \figref{fig:graphStruct}(a). Also, each node $v \in \mathcal{V}$  is associated with a 3-dimensional feature vector $\mathbf{x}_v = (\delta_x, \delta_y, \delta_z) \in \mathbb{R}^3$, representing the displacement of that node under externally applied forces. These displacement vectors are computed using nonlinear FE analysis and encode the spatial deformation responses of the tissue, implicitly reflecting local mechanical properties such as stiffness.

To define the edge set $\mathcal{E}$, we combine two types of edges: $\mathcal{E}_d$ and $\mathcal{E}_s$ such that $\mathcal{E} = \mathcal{E}_d \cup \mathcal{E}_s$. First, $\mathcal{E}_d$ encodes local spatial relationships based on Euclidean distance between nodes. Specifically, as shown in \figref{fig:graphStruct}(b), the nodes $v_A$ and $v_B$ are connected if and only if $|r_\text{A}-r_\text{B}|<0.003$ m, where $r_\text{A}$ and $r_\text{B}$ denote the position of node $v_A$ and $v_B$, respectively. These distance-based edges help the GNN learn local mechanical relationships by aggregating information from spatially neighboring nodes.

In addition to the distance-based graph construction, we also implement a \textbf{structured edge augmentation} approach and construct an additional edge set $\mathcal{E}_s$ to promote enhanced information transfer within the model (\figref{fig:graphStruct}(c)). Specifically, $\mathcal{E}_s$ is constructed by connecting a small subset of nodes in the breast surface and the cancer mass surface. As a result, each newly constructed edge in $\mathcal{E}_s$ directly links one node on the cancer surface with a node on the breast surface, enabling the model to effectively infer the deformation of the cancer nodes directly based on the information propagated from surface nodes via $\mathcal{E}_s$. This approach greatly alleviates the pressure of the GNN to preserve information from the surface nodes for multiple message-passing rounds. Without structured edge augmentation, message-passing in the GNN is limited to a distance of $0.003$ m multiplied by the number of GNN layers, which restricts the capability of the model to learn and spread information effectively.

\begin{figure*}[ht!]
    \centering
    \includegraphics[width=0.95\textwidth]{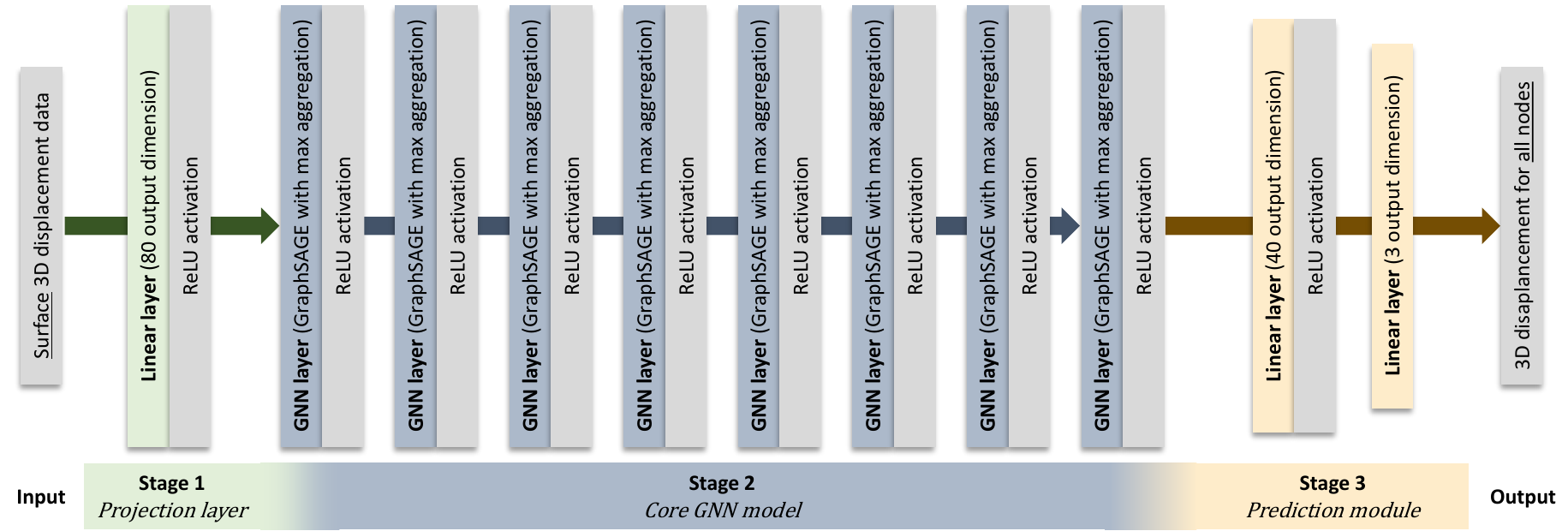}
    \caption{Multi-layer GNN-based architecture used in this study. The input surface data is first passed through a linear layer with ReLU activation, followed by 8 GNN layers (each with ReLU). The model output is then generated via a 2-layer MLP, effectively reducing dimensionality.}
    \label{fig:GNNmodel}
\end{figure*}

\subsection{GNN model architecture}
As we have acquired the graph data, we shift our focus to the GNN architecture used in our method. 

Our GNN model architecture is comprised of three major stages: the initial projection layer, followed by the sequence of multiple GNN layers, and finally a prediction module. In the remainder of this section, we will provide a detailed description of each steps in our architecture. The overall structure of our model is illustrated in \figref{fig:GNNmodel}.

\paragraph{Stage 1 (Projection layer).} In our initial stage, we employ a projection layer, which is implemented by a linear layer that takes $3$-dimensional vectors (\textit{i.e.}, displacement vector for each node) and outputs a $d$-dimensional vector ($d > 3$) followed by a ReLU layer for non-linearity. Although this process in itself does not introduce substantially new information, in a practical sense, the GNN layers in the next stage will not achieve its best performance without it. Intuitively, if we do not expand the dimension of the embedding vectors, the GNN will be forced to squeeze too much information from multiple nodes in just 3-dimensions during its message-passing layers. Therefore, by implementing the projection layer, we expand the dimensions of the embedding vector as early as possible.

\paragraph{Stage 2 (Core GNN model).} In this stage, a sequence of GNN layers will process both the embedding vectors from the previous stage containing the surface displacement information alongside the graph topology constructed in Section~\ref{subsection:graphconstruction}. The core mechanism of a GNN layer is based on the message-passing framework~\citep{Gilmer2017messagepassing}, where it involves two steps, \textit{i.e.}, \textit{aggregate} and \textit{update}. More specifically, for each node $v$ and its $d$-dimensional embedding vector $\mathbf{h}_v \in \mathbb{R}^d$, the GNN layer computes a new $d'$-dimensional embedding vector $\mathbf{h}'_v  \in \mathbb{R}^{d'}$ by the following two steps. 
\begin{enumerate}
    \item \textit{(Aggregate)} The layer collects relevant information from the neighbor nodes of $v$, denoted as $\mathcal{N}_v = \{(w, v) | (w, v) \in \mathcal E\}$.
    \item \textit{(Update)} Given the collected information from the aggregate step and $\mathbf{h}_v$, update the embedding vector to $\mathbf{h}'_v$.
\end{enumerate}
Intuitively, each message-passing layer propagates the information of a node to each immediate neighbors in $\mathcal{N}_v$. The specific details for each step (\textit{e.g.}, what information to collect and how to update) differ for different GNN architectures. In our implementation, we employ the message-passing layer from GraphSAGE~\citep{Hamilton2017SAGE} with max aggregation, which can be effectively expressed by the following mathematical form:

\begin{equation}\label{eq:gnnlayer}
    \mathbf{h}'_v = \mathbf{W}_1 \mathbf{h}_v + \mathbf{W}_2 \text{max}(\{ 
\mathbf{h}_w | w \in \mathcal{N}_v \}),
\end{equation}
where $\mathbf{W}_1, \mathbf{W}_2 \in \mathbb{R}^{d' \times d}$ are learnable weight matrices, and the $\text{max}(\mathcal{S})$ function accepts a set of vectors $\mathcal{S}$ as input and returns a vector of the element-wise maximum from the vectors in $\mathcal{S}$. In practice, we set $d$ and $d'$ to be the same across all layers unless stated otherwise.

\paragraph{Stage 3 (Prediction module).} After the model has performed multiple rounds of message passing, we finally employ the prediction module. The main objective is to process the output of the GNN back into a 3-dimensional displacement vector while preserving the relevant information as much as possible. To this end, we use a 2-layer MLP with ReLU activation for our implementation, where it reduces the dimension from $d$ to $d/2$, and finally down to $3$.

\paragraph{Default setting.} In our default setting, we stack 8 GNN layers described in~\eqref{eq:gnnlayer} for effective information propagation. We used MSELoss as the loss function, and AdamW~\citep{Loshchilov2019adamw} as the optimizer. Each GNN layer has 80 hidden channels, \textit{i.e.}, $d=80$. The number of training epochs was set to 500 based on empirical observations that validation performance converges around this point, balancing model fitting and generalization. We adopted this configuration as the default setting after conducting ablation studies on various architectural components, as detailed in Section~4.2.

\section{Results}
This section presents the sensitivity analysis of the model using a breast phantom and its performance evaluation on real patient data. All tests were conducted on a system with an AMD Ryzen 9 5950X 16-core processor (3.4 GHz), 128 GB of RAM, and an NVIDIA GeForce RTX 4080 GPU with 16 GB of VRAM.

\subsection{Evaluation metrics}
To assess the performance of the proposed GNN model, two evaluation metrics are employed: (1) RMSE for prediction accuracy of nodal displacement and (2) DSC for spatial overlap of the cancerous region.

To quantify the accuracy of the predicted displacement at each node, the RMSE is defined as:
\begin{equation}
\text{RMSE} = \sqrt{\frac{1}{N}\Sigma(\delta_\text{T}-\delta_\text{P})^2},
\end{equation}
where $\delta_\text{T}$ and  $\delta_\text{P}$ denote the target and predicted displacement values at each node, respectively, and $N$ represents the total number of nodes. For more detailed evaluation, RMSE is computed in two forms: (1) global RMSE is averaged over all nodes in the model to assess the overall displacement prediction accuracy and (2) cancer region-specific RMSE (referred to as cancer RMSE hereafter) is computed exclusively for the nodes corresponding to the cancerous region to evaluate the prediction accuracy in the clinically significant area.


To evaluate the volume conformity accuracy of the predicted cancerous region, DSC is utilized. Since the FE and GNN models are node-based, the target and predicted cancer regions are voxelized using the winding number algorithm \citep{HORMANN200113} at a resolution of 1 mm before computing DSC, which is defined as:
\begin{equation}
\text{DSC} = \frac{2|V_\text{T} \cap V_\text{P}|}{|V_\text{T}| + |V_\text{P}|}
\end{equation}
where $V_{\text{T}}$ and $V_{\text{P}}$ represent the sets of voxels classified as part of the cancerous region in the target and predicted data, respectively.

\subsection{Sensitivity analysis with phantom data}
We conduct a sensitivity analysis using P1 dataset to assess the influence of each component in our network model. By evaluating their effects on prediction accuracy, we aim to identify the key factors that contribute to the model's effectiveness and provide insights for future improvements.

\subsubsection{Effect of GNN layer architecture}
Table \ref{tab:GNNLayer} shows a comparison of prediction accuracy among different GNN architectures, including GCN~\citep{Kipf2017gcn}, GraphSAGE~\citep{Hamilton2017SAGE}, and GraphConv~\citep{morris2019graphconv}, with the MLP result as a reference model. While both GraphSAGE and GraphConv significantly outperform MLP, demonstrating the effectiveness of graph-based representations, GCN did not show improvement over the MLP baseline and exhibited the lowest accuracy among the tested GNN models. Among the GNN architectures, GraphSAGE achieved the lowest cancer RMSE of 0.093 mm, indicating its superior accuracy in predicting cancer locations. Although GraphConv also demonstrated comparable results, it showed slightly lower accuracy than GraphSAGE in terms of cancer RMSE, which is the clinically important metric. Based on these findings, GraphSAGE is selected as the primary architecture for further analysis.

\begin{table}[!htbp]
    \centering
    \caption{Prediction accuracy comparison across GNN architectures and MLP baseline. Configurations: 8 layers, 80 channels, and 100 structured edge augmentations.}
    \begin{tabular}{cccc}
    \toprule
        Architecture & Global RMSE & Cancer RMSE & DSC \\
        \midrule
        GCN & 1.185 & 1.586 & 0.847\\
        \rowcolor{lightgray} GraphSAGE & 0.932 & 0.093 & 0.992\\
        GraphConv & 0.857 & 0.099 & 0.992\\
        MLP & 2.454 & 0.279 & 0.835 \\ 
        \bottomrule
    \end{tabular}
    \label{tab:GNNLayer}
\end{table}

\subsubsection{Effect of layer depth}
Table \ref{tab:GNNDepth} presents the prediction accuracy depending on the GNN layer depth, ranging from 6 to 12 layers. The results show that increasing the number of layers does not consistently improve performance. The best results were achieved with 8 layers, where both global RMSE and cancer RMSE were the lowest (0.932 mm and 0.093 mm, respectively). Although a deeper network with 10 layers yielded slightly worse performance, the 6-layer model also demonstrated relatively low accuracy. Based on these results, an 8-layer GNN architecture is chosen for optimal performance in terms of prediction accuracy.

\begin{table}[!htbp]
    \centering
    \caption{Prediction accuracy comparison across different GNN layer depths. Configurations: GraphSAGE architecture, 80 channels, and 100 structured edge augmentations.}
    \begin{tabular}{cccc}
    \toprule
        Layers & Global RMSE & Cancer RMSE & DSC \\
        \midrule
        6 & 0.876 & 0.111 & 0.991\\
        \rowcolor{lightgray} 8 & 0.932 & 0.093 & 0.992\\
        10 & 1.045 & 0.099 & 0.992\\
        12 & 0.893 & 0.115 & 0.990\\
        \bottomrule
    \end{tabular}
    \label{tab:GNNDepth}
\end{table}

\subsubsection{Effect of number of channels} 
Table \ref{tab:GNNChannels} presents the prediction accuracy across different numbers of channels, ranging from 64 to 96. The results show that the highest accuracy in terms of cancer RMSE was achieved with 80 channels, where the cancer RMSE was the lowest (0.093 mm). Although the model with 96 channels demonstrated the best performance in terms of global RMSE, the cancer RMSE, which is considered as the primary metric in clinical applications, was lowest with the 80 channel model. Therefore, the model with 80 channels is selected for its optimal cancer RMSE performance.

\begin{table}[!htbp]
    \centering
    \caption{Prediction accuracy comparison across different number of channels. Configurations: GraphSAGE architecture, 8 layers, and 100 structured edge augmentations.}
    \begin{tabular}{cccc}
    \toprule
        Channels & Global RMSE & Cancer RMSE & DSC \\
        \midrule
        64 & 0.922 & 0.112 & 0.991\\
        72 & 1.008 & 0.112 & 0.991\\
        \rowcolor{lightgray} 80 & 0.932 & 0.093 & 0.992\\
        96 & 0.888 & 0.100 & 0.992\\
        \bottomrule
    \end{tabular}
    \label{tab:GNNChannels}
\end{table}

\subsubsection{Effect of structured edge augmentation}
Table \ref{tab:GNNEdgeAug} presents the prediction accuracy across different numbers of structured edge augmentations, ranging from 0 to 200. The results show that incorporating edge augmentation improved performance compared to the model without any augmentation (0 augmentations). The model with 100 augmented edges achieved the best cancer RMSE (0.093 mm), while the performance with 200 augmentations showed a slight increase in cancer RMSE (0.104 mm). Based on these results, the model with 100 structured augmented edges is selected for its optimal performance for further analysis.

\begin{table}[!htbp]
    \centering
    \caption{Prediction accuracy comparison across different number of structured edge augmentation. Configurations: GraphSAGE architecture, 8 layers, and 80 channels.}
    \begin{tabular}{cccc}
    \toprule
        Edges & Global RMSE & Cancer RMSE & DSC \\
        \midrule
        0 & 1.144 & 0.831 & 0.942\\
        \rowcolor{lightgray} 100 & 0.932 & 0.093 & 0.992\\
        200 & 0.893 & 0.104 & 0.991\\
        \bottomrule
    \end{tabular}
    \label{tab:GNNEdgeAug}
\end{table}

\subsubsection{Effect of jumping knowledge}
Jumping knowledge~\citep{Xu2018JumpingKnowledge} is a technique that utilizes all intermediate embedding vectors instead of the output of the final layer. Intuitively, this allows the model to utilize all intermediate embedding vectors, each of which contains information from different sets of neighboring nodes. Denoting $\mathbf{h}_v^{(p)}$ as the output vector of the $p$-th GNN layer of node $v$, jumping knowledge calculates $\mathbf{h}^{\text{JK}}_v = \text{max}(\{\mathbf{h}_v^{(1)}, \cdots, \mathbf{h}_v^{(p)}\})$, which is used as the input to the next module. Table \ref{tab:GNNJK} presents the prediction accuracy comparison between models depending on the use of jumping knowledge, with comparisons made for the 8 layer and 10 layer configurations. Overall, the 8 layer configuration showed slightly higher performance than the 10 layer model in all metrics. The use of jumping knowledge resulted in an improvement in global RMSE, which decreased from 0.932 mm to 0.875 mm. However, in terms of cancer RMSE, which is the clinically significant metric, a slight degradation in performance was observed, with the value increasing from 0.093 mm to 0.100 mm. Based on these results, the decision was made to exclude the use of jumping knowledge, as it did not provide a significant benefit in the clinically relevant cancer RMSE metric.

\begin{table}[!htbp]
    \centering
    \caption{Prediction accuracy comparison with and without jumping knowledge. Configurations: GraphSAGE architecture, 8 layers, 80 channels, and 100 structured edge augmentations.}
    \begin{tabular}{cccc}
    \toprule
        Layers & Global RMSE & Cancer RMSE & DSC \\
        \midrule
        \rowcolor{lightgray} 8 no JK & 0.932 & 0.093 & 0.992\\
        8 with JK & 0.875 & 0.100 & 0.992\\
        10 no JK & 1.045 & 0.099 & 0.992\\
        10 with JK & 0.928 & 0.102 & 0.991\\
        \bottomrule
    \end{tabular}
    \label{tab:GNNJK}
\end{table}

\subsection{Performance evaluation on patient dataset}
Based on the sensitivity analysis in Section 4.2, the following configurations were selected for further testing on datasets H1, H2 and H3: GraphSAGE was used as the layer architecture, the layer depth was set to 8, the number of channels was set to 80, structured edge augmentation was set to 100, and jumping knowledge was not used.

\begin{figure*}[ht!]
    \centering
    \includegraphics[width=0.95\textwidth]{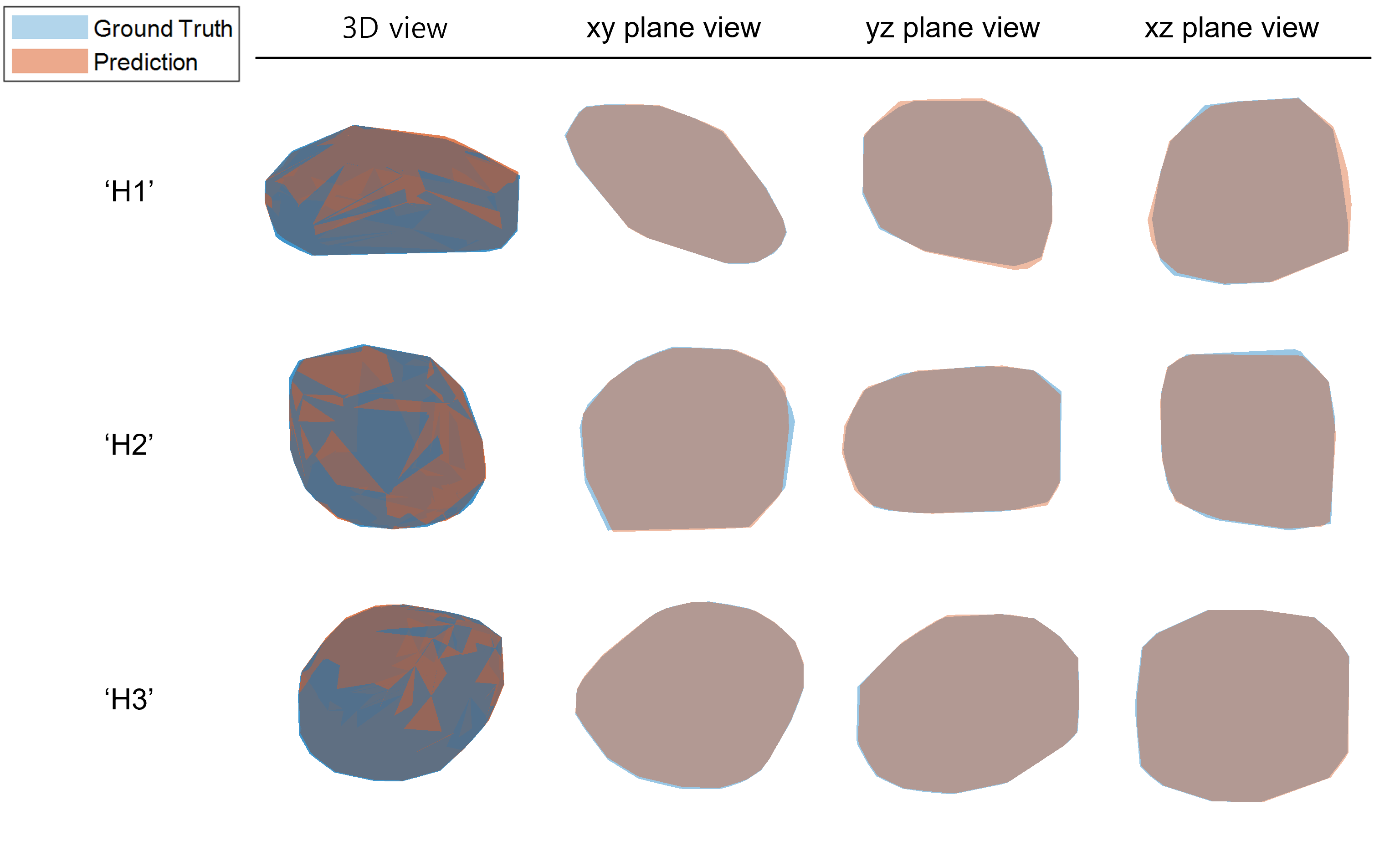}
    \caption{Exemplar deformed configuration of cancerous region in three patient datasets ('H1', 'H2', and 'H3'). For each dataset, the ground truth (blue) and GNN-predicted (red) deformed shapes are visualized in 3D, as well as in three 2D planes: axial (xy), coronal (yz), and sagittal (xz).}
    \label{fig:Example}
\end{figure*}

\figref{fig:Example} illustrates exemplar predictions of the deformed tumor region for three patient datasets (H1, H2, and H3). For each case, the GNN-predicted shape (red) is compared with the ground truth (blue) in both 3D and three orthogonal 2D planes (axial, coronal, and sagittal views). The close alignment between the predicted and actual deformed shapes confirms that the model successfully learned to capture the deformation characteristics specific to each patient.

\begin{figure*}[ht!]
    \centering
    \includegraphics[width=1.00\textwidth]{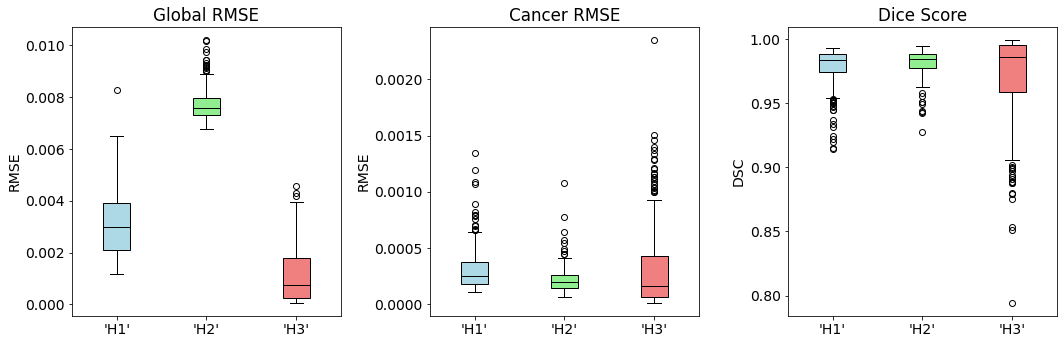}
    \caption{Boxplots of global RMSE, cancer RMSE, and DSC for three real patient datasets ('H1', 'H2', and 'H3'). The black horizontal line inside each box represents the median, while the box spans the interquartile range (IQR), from the first quartile (Q1) to the third quartile (Q3). The whiskers extend to the lower (Q1 – 1.5×IQR) and upper (Q3 + 1.5×IQR) bounds, indicating the range of non-outlier data points. Outliers beyond the whiskers are shown as individual points.}
    \label{fig:boxplot}
\end{figure*}

Table \ref{tab:AccEval} shows the quantitative performance of the GNN models for the three patient datasets (H1, H2, and H3). While performance variation is observed across the datasets, the models achieved high overall accuracy, with a mean global RMSE of 3.965 mm, mean cancer RMSE of 0.278 mm, and mean DSC of 0.977. \figref{fig:boxplot} presents boxplots of global RMSE, cancer RMSE, and DSC for each patient dataset.

Table \ref{tab:TimeEval} compares the average computational time required per data sample for conventional FE simulations and GNN inference across the three patient datasets. While the FE simulations took an average of 24.45 seconds per sample, the GNN model reduced the inference time to 5 milliseconds, corresponding to a speed-up of over 4,000 times. This significant reduction in computational cost highlights the model’s capacity for real-time deformation prediction, making it highly suitable for time-sensitive clinical applications.

\begin{table}[!htbp]
    \centering
    \caption{Quantitative evaluation of GNN accuracy on three patient datasets}
    \begin{tabular}{cccc}
    \toprule
        Datasets & Global RMSE & Cancer RMSE & DSC \\
        \midrule
        'H1' & 3.078 & 0.306 & 0.979 \\
        'H2' & 7.710 & 0.213 & 0.982\\
        'H3' & 1.108 & 0.315 & 0.971\\
        \midrule
        Mean & 3.965 & 0.278 & 0.977\\
        \bottomrule
    \end{tabular}
    \label{tab:AccEval}
\end{table}

\begin{table}[!htbp]
    \centering
    \caption{Comparison of finite element simulation time and GNN inference time across datasets (seconds per data sample)}
    \begin{tabular}{cccc}
    \toprule
        Datasets & FE simulation & GNN inference & Speed-up \\
        \midrule
        'H1' & 27.51 s & 0.005134 s & 5358$\times$\\
        'H2' & 28.57 s & 0.007465 s & 3827$\times$\\
        'H3' & 17.28 s & 0.005009 s & 3450$\times$\\
        \midrule
        Mean & 24.45 s & 0.005869 s & 4166$\times$\\
        \bottomrule
    \end{tabular}
    \label{tab:TimeEval}
\end{table}

\section{Discussion}
For out-bore MRI-guided breast biopsy, accurately predicting tumor displacement due to breast deformation is crucial for ensuring precise needle placement. However, existing computational models struggle to achieve real-time performance due to their high computational cost, limiting practical clinical application. To address this challenge, our study leverages the predictive efficiency of GNNs trained on FE analysis results to model breast deformation in real time. By incorporating MRI-derived structural information and surface displacement data, our approach enables accurate tumor localization without requiring direct force measurements. Notably, validation using both breast phantom and real patient datasets demonstrated sub-millimeter accuracy for predicting cancer displacement within real-time, highlighting its potential for clinical integration.

The results demonstrated that prediction accuracy varied depending on the choice of GNN architecture. This performance difference can be attributed to each model’s ability to capture complex and spatially heterogeneous deformation patterns. In particular, GNN architectures that process the self information and the aggregated information separately—beyond the simple averaging used in GCN—were able to better represent localized and directional tissue deformation. Both GraphSAGE and GraphConv exhibited improved performance by the separate processing strategies, mitigating the loss of self-information during multiple message-passing steps and eventually leading to higher accuracy in predicting tumor displacement. These findings highlight the importance of selecting GNN architectures with advanced aggregation strategies for accurately modeling complex biomechanical behavior.

Unlike many typical GNN applications, where increasing the number of layers often leads to other problems such as over-smoothing~\citep{Li2018oversmoothing,NT2019oversmoothing,OonoS20oversmoothing}, our study found that using a relatively deep eight-layer architecture provided the best accuracy. This result can be interpreted in the context of information propagation distance. Predicting tumor displacement due to complex tissue deformation requires capturing long-range interactions between boundary conditions and interior regions of the breast tissue. A deeper network allows information to travel across multiple hops, effectively integrating these distant dependencies. Furthermore, the structured graph design, combined with structured edge augmentations, may have facilitated stable information flow without losing feature distinctiveness. These observations highlight that, in physics-based deformation modeling, sufficient network depth is crucial to ensuring that global boundary information is transmitted and reflected in local displacement predictions.

While the study by \citep{Salehi2022PhysGNN} demonstrated the benefits of jumping knowledge in improving information aggregation across layers, our results indicate that jumping knowledge does not provide a significant advantage in our study. This difference can be attributed to the structured edge augmentation employed in our model, which effectively enhances connectivity and facilitates information propagation. By strategically augmenting edges, the network inherently mitigates the need for long-range feature aggregation through jumping knowledge, as critical structural relationships are already well-preserved. These findings suggest that the necessity of jumping knowledge may be highly dependent on the underlying graph topology, and in cases where connectivity is sufficiently optimized, alternative mechanisms such as edge augmentation may serve as a more efficient means of improving representation learning.

The significant performance improvement observed with structured edge augmentation highlights the crucial role of enhanced connectivity in the GNN model. Unlike task-agnostic augmentation strategies, our method selectively introduces edges that directly connect surface nodes to the cancer mass, embedding physics-guided, task-specific knowledge into the graph and representing a novel contribution in this context. These structured edges create shortcut pathways that allow critical mechanical information to propagate efficiently in just one or two message-passing steps. This reduces signal attenuation and oversmoothing that can occur when relying solely on local mesh adjacency, particularly in highly nonlinear deformation scenarios. Consequently, the network forms richer and more physically meaningful feature representations, improving predictive accuracy while maintaining computational efficiency. Overall, these findings underscore the importance of carefully designing graph topology to optimize learning in physics-guided neural networks, demonstrating both the novelty of the edge augmentation approach and its effectiveness in enhancing message passing.

While structured edge augmentation substantially improves prediction accuracy in the lesion region, as evidenced by the significant reduction in cancer RMSE (e.g., from 0.831 to 0.093 in Table \ref{tab:GNNEdgeAug}), the improvement in global RMSE is comparatively modest. This discrepancy stems from the selective nature of our edge augmentation strategy, which introduces direct connections only between surface nodes and the lesion region. Such targeted connectivity facilitates efficient message passing specifically to and from the cancer mass, enabling the model to form expressive and localized feature representations that are critical for accurate lesion-specific prediction. In contrast, regions distant from the lesion rely solely on the baseline graph construction, where edges are formed between nearby nodes based on a fixed Euclidean distance threshold (0.003 m). While this preserves local geometric structure, it creates a trade-off between sparse connectivity and oversmoothing: a low threshold limits long-range interactions, whereas a high threshold can lead to excessive edge density, diminishing model expressiveness. To resolve this, we adopted a hybrid strategy—maintaining a conservative global threshold while selectively enhancing connectivity in clinically relevant areas through structured edges. This design reflects our emphasis on lesion-focused prediction, but it also explains why the model’s performance gains are more pronounced in the cancer region than across the full domain. Future work may explore more globally adaptive or deformation-aware edge augmentation strategies to further reduce global RMSE without compromising local prediction fidelity near the lesion.

While the proposed modeling framework demonstrates strong predictive performance, its generalizability is constrained by assumptions in the FE simulations. Specifically, fixed hyperelastic material properties were used for both normal and tumor tissues, based on values reported in prior literature. However, tissue mechanical characteristics can vary significantly between individuals, which limits the ability to capture patient-specific behavior. To address this limitation, future work must first overcome the challenge of non-invasively estimating the mechanical properties of both normal and tumor tissues in vivo. Without such techniques, personalized and physically accurate modeling remains difficult. Additionally, the current FE model simplifies breast anatomy into two homogeneous tissue types, thereby neglecting the heterogeneous and multi-component structure of real breast tissue. Incorporating more detailed anatomical structures and patient-specific material characterization methods would further enhance both the realism and clinical applicability of the proposed approach.

Although the proposed model effectively captures deformation induced by external mechanical forces during biopsy procedures, it does not currently account for physiological motions such as respiration or muscle contraction. While rigid-body registration between the MR image and the physical space compensates for global patient motion, localized non-rigid deformations resulting from breathing or muscle activity are not explicitly modeled. This simplification is based on the assumption that the dominant source of deformation originates from procedural interactions. Nevertheless, the inclusion of dynamic physiological factors could further improve prediction accuracy and patient-specific fidelity. Incorporating such effects remains an important direction for future research.

One limitation of the current approach is that the network model is trained individually for each patient, enabling highly customized and accurate predictions but requiring additional time to prepare and optimize the network before clinical use. This patient-specific training process may limit its applicability in certain clinical scenarios where rapid deployment is required. However, the envisioned clinical workflow in this study is designed to accommodate this requirement: once a diagnostic MRI is acquired, the full modeling pipeline—including finite element model construction, simulation-based dataset generation, and GNN training—can be completed within one day. This allows the trained model to be used for an indirect MRI-guided biopsy procedure scheduled for the following day, offering both feasibility and high predictive accuracy in a practical clinical timeline. Moreover, since the model is trained exclusively on each patient’s MRI-derived data, the risk of bias due to limited anatomical diversity is inherently minimized. This patient-specific workflow thus provides a balanced solution between individualized precision and clinical practicality.

To improve scalability, future work could explore inductive learning or transfer learning approaches. Although the diversity in breast geometry and tumor characteristics presents challenges for direct generalization, we observe that the overall topology of breast deformation remains relatively consistent across patients. Leveraging this structural consistency may enable the development of topology-aware representations, providing a promising pathway toward generalizable models that require less retraining. Such approaches could significantly reduce model preparation time while maintaining prediction performance, thereby improving the practicality of the proposed method in broader clinical applications.

While the current study demonstrates high accuracy under ideal conditions, practical deployment in clinical settings introduces additional challenges. Accurate real-time surface displacement acquisition using depth-sensing cameras may be affected by noise, occlusions (e.g., surgical tools or hands), or missing data. To ensure the robustness of the proposed method under such conditions, future work will include uncertainty quantification (UQ) and sensitivity analyses with synthetic noise, as well as experiments using real-time sensor integration. These efforts aim to evaluate the system’s performance and resilience in realistic settings and are critical steps toward clinical translation.

For the proposed model to be translated into clinical practice, future work should consider its integration into the biopsy workflow, including real-time sensing, visualization, and human–system interaction. One promising direction is to link the GNN model with depth-sensing cameras that capture intraoperative surface displacement, enabling real-time tracking of tumor movement during biopsy procedures. These predicted displacements can be transmitted to an augmented or virtual reality (AR/VR) device to assist clinicians in navigation by visually overlaying the predicted tumor position on the patient's anatomy. Such systems can be feasibly implemented using commercially available AR devices that receive displacement data from a nearby personal computer via wireless connection, with minimal additional hardware. Future studies should also investigate the usability and ergonomics of this system in realistic clinical environments, as well as assess the impact of user feedback and potential workflow disruptions. Importantly, processes such as surface displacement acquisition and wireless AR data transfer are generally achieved within a fraction of a second on standard hardware, suggesting that the entire integrated pipeline—including GNN inference—can achieve end-to-end latency well below one second. Establishing such a real-time, integrated pipeline will be essential for realizing the translational potential of the proposed framework.

In clinical practice, particularly in indirect MRI-guided biopsy workflows, an error margin of up to approximately 5 mm is generally regarded as acceptable for safe and effective needle targeting, especially when combined with optical tracking systems \citep{krucker2007electromagnetic}. In our experiments, the proposed deformation-aware model achieved cancer localization errors well below this threshold—for example, 0.278 mm as shown in Table 6—demonstrating strong potential for high-precision targeting. While this study focuses on the core deformation prediction model, we acknowledge that additional sources of error may arise in a complete workflow, such as depth camera-based surface measurements and MRI-to-patient space registration. However, based on typical performance reported in prior studies, these components usually contribute errors on the order of 1–2 mm. Even when such uncertainties are considered, the cumulative targeting error is expected to remain within the clinically accepted 5 mm range. These findings support the feasibility of translating our method into real-time guidance applications for needle-based interventions.

Although this study focused on indirect MRI-guided breast biopsy, the proposed framework holds significant potential for broader applications. Needle-based interventions \citep{shin2024RFA}, such as radiofrequency (RF) ablation or microwave (MW) ablation, similarly require precise needle placement under complex tissue deformation. In these procedures, accurately predicting target displacement due to mechanical deformation remains a critical challenge. The GNN-based deformation prediction model developed in this study, which leverages structural imaging data and surface displacements, can be extended to these applications without substantial modifications. By enabling real-time prediction of target locations, this approach has the potential to enhance procedural accuracy and safety across a wide range of needle insertion therapies.

\section{Conclusion}
This study presents a novel approach for real-time breast tumor deformation prediction, utilizing a GNN trained on FE simulation results. By incorporating MRI-derived breast and tumor structures into the FE model, we were able to simulate patient-specific deformation behaviors. The GNN model, using surface displacement data as an input feature, demonstrated high predictive accuracy and real-time capability, which was validated through phantom and real patient datasets. The proposed technique, when integrated with indirect MRI-guided biopsy, holds significant potential for enhancing the precision and efficiency of breast biopsy procedures. Future research will focus on extending the framework toward clinical integration and real-time system implementation, improving model generalization through transfer learning, and incorporating non-rigid motion components such as respiratory deformation to ensure robustness under dynamic clinical conditions.

\section{Declaration of competing interest}
The authors declare that they have no known competing financial interests or personal relationships that could have appeared to influence the work reported in this paper.

\section{Acknowledgement}
The work was supported by the National Research Foundation of Korea (NRF) grant funded by the Korean government (MSIT) (No. RS-2023-00220762 and No. RS-2024-00335185).
\bibliographystyle{elsarticle-num-names}
\bibliography{references}

\end{document}